# Recurrent Transform Learning

Megha Gupta, and Angshul Majumdar, *Senior Member, IEEE*

*Abstract*—The objective of this work is to improve the accuracy of building demand forecasting. This is a more challenging task than grid level forecasting. For the said purpose, we develop a new technique called recurrent transform learning (RTL). Two versions are proposed. The first one (RTL) is unsupervised; this is used as a feature extraction tool that is further fed into a regression model. The second formulation embeds regression into the RTL framework leading to regressing recurrent transform learning ($R^2TL$). Forecasting experiments have been carried out on three popular publicly available datasets. Both of our proposed techniques yield results superior to the state-of-the-art like long short term memory network, echo state network and sparse coding regression.

## I. INTRODUCTION

THE importance of electrical load forecasting is well known. The issue has gained even more significance with the advent of smartgrids, microgrids and smart buildings. An excellent review on this topic can be found in [1]. While the aforesaid review is of technical nature, there are other review articles on the topic of demand response delving into the financial consequences [2, 3]. This work addresses the technical problem of forecasting demand. The financial aspects of the problem will not be discussed; the interested reader may peruse the aforesaid review articles.

Broadly speaking there are three approaches to load forecasting:
1. Linear regression [11-16]
2. Dynamical model [17-26]
3. Non-linear regression [27-37]

There are several interesting techniques that are employed for forecasting, which do not fit into any of the aforesaid major categories [38-43]. Moreover, in recent times, the success of deep learning in other areas of data analysis motivated researchers in this area to employ such models in load forecasting.

It must be noted that linear / non-linear regression models (not counting autoregressive moving average class of regression) are static in nature. They cannot inherently model the dynamical / time-varying nature of the problem. They handle regression by windowing technique. Recent studies in neural networks addresses this problem by incorporating feedback into the system. This led to the development of recurrent neural network (RNN) and its variants like echo state network (ESN) and long short-term memory networks (LSTM). These techniques combine the power of non-linearity with the ability to incorporate dynamic behavior of the problem. One can even build deeper architectures by using them one after the other. In recent times, such techniques are gaining popularity for demand forecasting [4-6].

However, it must be remembered that LSTM / ESN / RNN cannot directly predict. They can only extract features from the sequence. These features must be fed to a fully connected network for further analysis. The entire network has to be solved via backpropagation through time. This brings us to our proposed model. Our method can be both unsupervised and supervised. To the best of our knowledge, the unsupervised version of our work is the only approach that can model dynamic sequences in an unsupervised manner, given the neural network class of techniques. Our supervised version, provides a unified pipeline for inferring from sequences.

In this work, we propose a new approach to dynamical modeling. It is based on transform learning [7-9] – the analysis equivalent of dictionary learning [10]. The standard transform learning is equivalent to a feedforward neural network (as will be explained later). We incorporate the ability to model memory into the formulation by feeding back the previous output to the input. This is a standard approach in any dynamical model, e.g. RNN, Kalman Filter, Hidden Markov Model etc. Since, our model is loosely based on the neural network interpretation of transform learning, we have named it recurrent transform learning. As our goal is to predict the demand; and this is best modeled as a regression problem (because the outputs are real valued), we add a regression node at the output of our otherwise unsupervised recurrent transform learning model. This leads to our supervised regressing recurrent transform learning formulation.

Such jointly learnt formulations are known to yield better results than piecemeal techniques. This will be empirically verified when we compare our method with all state-of-the-art load forecasting techniques. We show that our model improves significantly over the state-of-the-art.

In this work the focus will be on load forecasting at the building level. This is an emerging application area and many recent studies are focused on this. Most prior studies were based on grid level forecasting; this was a much easier problem. The fluctuations at the building level gets ironed out at the grid level rendering highly accurate forecasting a relatively simple task. But each building being different, forecasting at the building level is a challenging task.

We believe that, such methods will be useful when alternate sources of energy will be available to the consumer [44]. For example, assume that the household is powered both by the electric grid as well as by solar energy; and that the house is able to forecast its next day's demand fairly accurately. Say that the household comes to know that the next day will be rainy, which means that production of solar energy will be low. In that case, the household may want to run heavy electricals such as washing machines and dishwashers today in order to prevent drawing too much from the grid (for which they have to pay) tomorrow.

In short there are two main contributions of the paper:

1. Developing a new machine learning / signal processing technique for unsupervised and supervised dynamical modeling.

2. Application of the proposed dynamical model to improve upon the state-of-the-art in short term load forecasting.

## II. RECURRENT TRANSFORM LEARNING

### A. Transform Learning

A transform analyses the data so as to generate the coefficients. Mathematically this is expressed as,

$$TX = Z \tag{5}$$

Here $T$ is the transform, $X$ is the data and $Z$ the corresponding coefficients. The data $X$ is organized as features along the rows and the samples along the columns. The number of transform basis used is decided by the user; this also defines the dimensions of the coefficients $Z$.

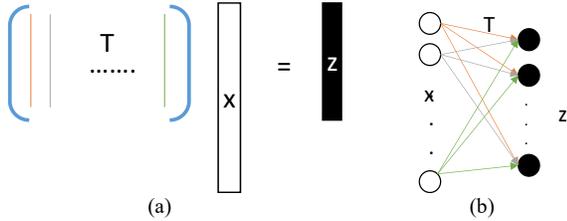

Fig. 1. (a) Transform Learning. (b) Neural Network Interpretation.

The model (5) can be interpreted in two ways. So far signal processing researchers have looked at it as an analysis basis – this is depicted in Fig. 1(a). It can alternately be depicted as an unsupervised feed forward neural network, as shown in Fig. 1(b). Instead of looking at the rows of the transform matrix as basis, one can think of them as connections between the input nodes and the representation nodes. This is akin to a feedforward unsupervised neural network.

The following transform learning formulation was proposed in [7, 8] –

$$\min_{T,Z} \|TX - Z\|_F^2 + \lambda\left(\|T\|_F^2 - \log\det T\right) + \mu\|Z\|_1 \tag{6}$$

The factor $-\log\det T$ imposes a full rank on the learned transform; this prevents the degenerate solution ($T=0, Z=0$). The additional penalty $\|T\|_F^2$ is to balance scale; without this $-\log\det T$ can keep on increasing producing degenerate results in the other extreme. Note that the sparsity penalty on the coefficients is not mandatory; it is essential only for solving inverse problems but does not carry any meaning, apart from being just a regularizer for machine learning tasks.

In [7, 8], an alternating minimization approach was proposed to solve the transform learning problem. This is given by –

$$Z \leftarrow \min_Z \|TX - Z\|_F^2 + \mu\|Z\|_1 \tag{7a}$$

$$T \leftarrow \min_T \|TX - Z\|_F^2 + \lambda\left(\varepsilon\|T\|_F^2 - \log\det T\right) \tag{7b}$$

Updating the coefficients (7a) is straightforward. It can be updated via one step of soft thresholding. This is expressed as,

$$Z \leftarrow signum(TX) \cdot \max\left(0, abs(TX) - \mu\right) \tag{8}$$

Here $\odot$ indicates element-wise product.

If the sparsity penalty is dropped, the update is simply $TX=Z$.

In the initial paper on transform learning [7], a non-linear conjugate gradient based technique was proposed to solve the transform update. In the more refined version [8], with some linear algebraic tricks they were able to show that a closed form update exists for the transform.

$$XX^T + \lambda\varepsilon I = LL^T \text{ (Cholesky decomposition)} \tag{9a}$$

$$L^{-1}XZ^T = USV^T \text{ (SVD)} \tag{9b}$$

$$T = 0.5V\left(S + (S^2 + 2\lambda I)^{1/2}\right)Q^T L^{-1} \tag{9c}$$

The first step is to compute the Cholesky decomposition; the decomposition exists since $XX^T + \lambda\varepsilon I$ is symmetric positive definite. The next step is to compute the full singular value decomposition (SVD). The final step is the update step. The proof for convergence of such an update algorithm can be found in [9].

### B. Proposed Technique

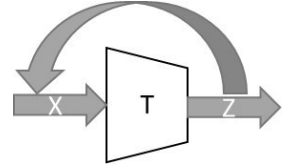

Fig. 2. Recurrent Transform Learning

The schematic diagram of our proposed formulation is shown in Fig. 2. It is a feedback system, the current output is dependent on the current input and the previous output. This is a standard approach to model any dynamical system; it has been used in Kalman Filters to Hidden Markov Models to Recurrent Neural Networks. Mathematically our formulation is expressed as,

$$T\begin{bmatrix} x_t \\ z_{t-1} \end{bmatrix} = z_t \tag{10}$$

In the context of demand forecasting, the inputs ($x_t$) are the loads from previous time instants over a window, weather data and any other information deemed relevant. The variable $z_t$ is the representation at the $t^{th}$ instant and $z_{t-1}$ is the representation from the previous instance that is being fed back along with the input at the current instant to model memory in the system. The columns are ordered by time.

The learning is expressed as,

$$\min_{T,z_t} \sum_t \left\| T\begin{bmatrix} x_t \\ z_{t-1} \end{bmatrix} - z_t \right\|_F^2 + \lambda\left(\|T\|_F^2 - \log\det T\right) \tag{11}$$

Note that we have dropped the sparsity penalty.

The sub-problems for solving (11) can be expressed as,

$$\min_T \left\| T\begin{bmatrix} X \\ Z \end{bmatrix} - Z \right\|_F^2 + \lambda\left(\|T\|_F^2 - \log\det T\right) \tag{12a}$$

$$\min_{z_t} \sum_{t=1}^{T} \left\| T \begin{bmatrix} x_t \\ z_{t-1} \end{bmatrix} - z_t \right\|_F^2 \quad (12b)$$

For updating the transform, we assume that the coefficients are constant. Therefore updating (12a) is exactly the same as (9).

Updates for the coefficients are more involved. We can express (10) in the following fashion –

$$T_1 x_t + T_2 z_{t-1} = z_t$$

Now we define a matrix $D_t$ for all $t$'s as follows,

$$D_t = I - TI_S \quad (13)$$

where $I_S$ is a shifted identity matrix and $I$ is an identity matrix. This allows expressing (12) as,

$$x = Dz \quad (14)$$

where $x = \begin{bmatrix} Tx_1 \\ ... \\ Tx_T \end{bmatrix}$, $D = [D_1 | ... | D_T]$ and $z = \begin{bmatrix} z_1 \\ ... \\ z_T \end{bmatrix}$.

Therefore solving z from (12b) turns out to be a simple linear inverse problem of the form,

$$\min_z \|x - Dz\|_2^2 \quad (15)$$

This has a closed form solution.

RTL Algorithm

---
Initialize T and $z_0$: Compute SVD of X, i.e. $X = USV^T$.
$T = U(:,1:NumOfBasis)^T$ and $z_0 = 0$.
Repeat Until Convergence or specified number of iterations

$$T \leftarrow \min_T \left\| T \begin{bmatrix} X \\ Z \end{bmatrix} - Z \right\|_F^2 + \lambda \left( \|T\|_F^2 - \log \det T \right)$$

$$z \leftarrow \min_z \|x - Dz\|_2^2$$
---

Using this basic formulation for recurrent transform learning (RTL) we can learn time varying features in an unsupervised fashion. This can be fed into a regression framework for load prediction. To the best of our knowledge this is the only work that can learn unsupervised features from a neural network class of models for dynamic sequences.

However, since our final goal is regression, a better approach would be to incorporate the regression into the RTL formulation – this would make the formulation supervised.

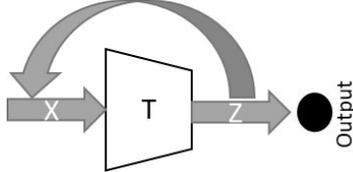

Fig. 3. Regressing Recurrent Transform Learning

The schematic diagram of regressing recurrent transform learning (R²TL) is shown in Fig. 3. The first part remains the same as (10). In the second part, we add the load ($l_t$) at the $t^{th}$ instance as the supervision output. Since the output is a real number, it turns out to be a regression model. This is modeled as,

$$l_t = w^T z_t \quad (16)$$

where $l_t$ is the regression output, i.e. the power consumption at a later point and $w$ is the vector of regression weights.

The joint learning model for R²TL can be expressed as follows,

$$\min_{T, z_t, w} \sum_t \left\| T \begin{bmatrix} x_t \\ z_{t-1} \end{bmatrix} - z_t \right\|_F^2 + \mu \|l_t - w^T z_t\|_2^2 + \gamma \|w\|_2^2$$
$$+ \lambda \left( \|T\|_F^2 - \log \det T \right) \quad (17)$$

Here the term $\|l_t - w^T z_t\|_2^2 + \gamma \|w\|_2^2$ corresponds to the regression formulation. We have used a ridge type penalty on the regression weights.

The complete formulation (17) can be segregated into the following updates:

$$\min_T \left\| T \begin{bmatrix} X \\ Z \end{bmatrix} - Z \right\|_F^2 + \lambda \left( \|T\|_F^2 - \log \det T \right) \quad (18a)$$

$$\min_{z_t} \sum_{t=1}^{T} \left\| T \begin{bmatrix} x_t \\ z_{t-1} \end{bmatrix} - z_t \right\|_F^2 + \mu \|l_t - w^T z_t\|_2^2 \quad (18b)$$

$$\min_w \mu \sum_{t=1}^{T} \|l_t - w^T z_t\|_2^2 + \gamma \|w\|_2^2 \quad (18c)$$

The update for (18a) remains the same as before (9).

The update for the coefficient changes slightly from the RTL formulation. We can express (18b) as follows,

$$\min_z \|x - Dz\|_2^2 + \mu \|l - Wz\| \quad (19)$$

Here $W = \begin{bmatrix} w^T & 0 & ... & 0 \\ 0 & w^T & ... & 0 \\ ... & ... & ... & ... \\ 0 & 0 & ... & w^T \end{bmatrix}$ and $l = \begin{bmatrix} l_1 \\ ... \\ l_T \end{bmatrix}$. The rest of the

symbols carry the same meaning as before. The solution to (19) is a simple since it is a least squares problem with known analytic formula.

The final step is to solve (18c) for updating the regression weights. This too is a least square problem having a closed form solution.

R²TL Algorithm

---
Initialize T and Z: Compute SVD of X, i.e. $X = USV^T$.
$T = U(:,1:NumOfBasis)^T$ and $Z = TX$.
Repeat Until Convergence or specified number of iterations

$$T \leftarrow \min_T \left\| T \begin{bmatrix} X \\ Z \end{bmatrix} - Z \right\|_F^2 + \lambda \left( \|T\|_F^2 - \log \det T \right)$$

$$z \leftarrow \min_z \|x - Dz\|_2^2 + \mu \|l - Wz\|$$

$$w \leftarrow \min_w \mu \|l - w^T z\|_2^2 + \gamma \|w\|_2^2$$
---

Both the RTL and the R²TL converges within 20 iterations. The transform learning based formulations are not convex, hence there is no guarantee of global convergence. We stop the iterations when the change in objective function in subsequent iterations is less than some tolerance.

This concludes the training phase for our formulations. For testing we will have the load information for the current window as well as weather and any other relevant information; our objective will be to predict the load for the next instant (day). For this, we need to generate the coefficients from the input data. This is achieved by solving (10) where $T$ is already learnt in the training phase. For both the RTL and the $R^2$TL algorithms, the testing phase remains the same; the coefficients are obtained by solving (12b). As mentioned, before this has a closed form solution. Once the coefficients are obtained, it is used in the pre-learnt regression model (a separate one for RTL and in-built one for $R^2$TL) to predict the load.

*C. Complexity Analysis*

Both the algorithms are iterative in nature; therefore, one can only give the computational complexity per iteration. For the RTL algorithm there are two sub-problems. The complexity of updating the transform is dominated by the cost of computing the singular value decomposition; this is given by $O(n^3)$. The complexity of updating the coefficients is given by computation of pseudoinverse; its complexity is $O(n^w)$ where $w<2.37$ and is conjectured to be 2. Therefore, the overall complexity is $O(n^3+n^2)$. For the $R^2$TL algorithm there is one additional step, that of updating the regression weights. This too has a solution via the pseudo-inverse and hence the overall complexity remains the same as before.

*D. Comparison*

In our proposal we build in memory into the transform learning formulation. This idea remains the same in all dynamical models – Kalman Filter (and its non-linear versions), Hidden Markov Model, recurrent neural network (and its variants). The similarity between our proposed work and RNN ends here. Note that the schematic diagram based on neural network is given just for the purpose of visualization. The model for RNN and our proposed one is not the same.

The crucial difference between RNNs and our proposed RTL is that, our method can be unsupervised, thereby allowing greater applicability. RNNs on the other hand have to be supervised. This brings to the second major difference – the training approach. Since RNNs have output, backpropagation through time can be applied. RTL on the other hand requires solution via more sophisticated optimization techniques since there is no output to backpropagate from.

## III. EXPERIMENTAL EVALUATION

*A. Datasets*

The REDD dataset[1] is a moderate size publicly available dataset. The dataset consists of power consumption signals from six different houses, where for each house, the whole electricity consumption as well as electricity consumptions of about twenty different devices are recorded. The device level information is required for energy disaggregation; that is not the goal of this work and hence will not be used. We will only use the data from the mains (total consumption). The signals from each house are collected over a period of two weeks with a high frequency sampling rate of 15kHz. To prepare training and testing data, aggregated and sub-metered data are averaged over a time period of 1 hour. In the standard evaluation protocol, the $5^{th}$ house is omitted since it has 80.85% of missing data. Data from the remaining 5 houses have been collected from between 17 to 26 days.

The Pecan Street dataset is obtained via the NILMTK[2]. It contains 1 minute circuit level and building level electricity data from 240 houses. The data set contains per minute readings from 18 different devices; however the device level information is used for energy disaggregation and will not be used in this work. We will only make use of the aggregate power data. The entire dataset contains more than 3 years of data, but make use of the subset provided by NILMTK.

The third dataset (IWAE) used in this work is from New Delhi, India[3]. The data was collected in a three storey building in Delhi, India, spanning 73 days from May-August 2013. This dataset too contains aggregate and sub-metered data for different appliances. But since our goal is not energy disaggregation we will not use the appliance level information. This dataset is available via the NILMTK as well.

For all the datasets, we collected the corresponding hourly weather (temperature and humidity) information at the city level. It is known that using the weather information improves prediction [43]. These values (arranged as a vector) were appended with the power consumption values and served as inputs to the algorithm.

*B. Comparative Methods*

Most recent studies in forecasting are based on the recurrent neural network (RNN) model. This has already been discussed (see Fig. 1). Here, the output of the current stage is a function of the current input and the output of the previous stage.

We carry out comparison with some state-of-the-art RNN based techniques in load forecasting. The first one is the Echo State Network (ESN) [34] – this is a variety of RNN. The configuration used in this work is borrowed from the prior study. They attained this using genetic algorithm based tuning. The number of principal components used is 5 (also from [31]).

The second method compared against is the deep long term memory network (LSTM) [4]; this too is a variant of the RNN. For the LSTM, we use a two layer architecture with 50 hidden units. This has shown to yield the best results in [4]; we tried other configurations but could not improve upon this configuration; the results reported in this paper using LSTM will be based on the said architecture.

The third technique is the sparse coding (SC) based regression [16]. In this technique, given the inputs, a dictionary and sparse coefficients are learnt by minimizing the Euclidean norm. The obtained sparse coefficients (codes) are then used in a standard regression framework for analysis.

---
[1] http://redd.csail.mit.edu/
[2] https://github.com/nilmtk/nilmtk
[3] http://iawe.github.io/

Here a dictionary of size 512 is used; the same has been proposed in [16]. We tried other configurations but could not improve. A separate regression step is used in [16]. It has been observed that the simple ridge regression yields consistently the best results; so we will use the same here.

We have also carried out comparison with traditional forecasting models like ARIMA approach, basic transform learning (in the same fashion as SC); i.e. transform learning is used to generate the coefficients which are then used by ridge regression for prediction and benchmark from [45]. However, the results were considerably poor compared to more recent ones used here; hence we are not showing these in the paper. These results can be found in the appendix

*C. Results*

For all the datasets we address the problem of one-day-ahead building load forecasting. This is a typical short-term load forecasting problem. The input consists of load, temperature and humidity data from previous days; we varied this window from past two to past seven days. We simply concatenated the power, temperature and humidity in one long input vector ($x_t$). The output consists of the total load for the next day.

For our proposed work, we have used two variants. The first one is RTL, where the feature is obtained separately and is fed into a ridge regression for prediction and the second one is $R^2TL$ where the regression is in-built. For both the formulations, the number of basis / atoms is kept at 50% of the input size. Note that the size of the input is not fixed, it varies with the window size (two to seven days). We tried varying the number of atoms but did not get any better results. With too few atoms, they are not able to capture the variability in the data and with too many atoms, they overfit. Both reduces the overall forecasting performance.

Our formulation requires specification of certain parameters. In RTL we only need $\lambda$ (11) and for $R^2TL$ we need $\lambda$, $\mu$ and $\gamma$ (17). These were fixed via the greedy L-curve technique. In this, $\gamma$ and $\mu$ are first set to 0 and $\lambda$ is tuned via the L-curve method; the obtained value is $\lambda=0.1$ (note that for RTL, this is the same as the standard L-curve since it has only one parameter to tune). The $R^2TL$ formulation also requires specifying $\mu$ and $\gamma$. The parameter $\mu$ controls the relative importance of the feature extraction terms and the regression term. Here we give equal importance to both; there we fix $\mu=1$. The parameter $\gamma$ is for the ridge penalty. We tune it by the greedy L-curve method where we used fixed values for $\lambda$ and $\mu$ (mentioned before); the value we obtain is 0.05.

In this work, we follow an experimental protocol similar to [16]. One half of the data (for each building) is used for training, the remaining half is used for testing. For tuning the parameters, all the algorithms used 5 fold cross validation using the training set. Evaluation is carried out in terms of three metrics – Mean Absolute Error (MAE), Root Mean Squared Error (RMSE) and Mean Absolute Percentage Error (MAPE).

In the first set of experiments, we compare our proposed technique with the ESN, LSTM and SC. The results for REDD, Pecan Street and IWAE are shown in Tables I, II and III respectively. We report the mean metrics for the REDD and Pecan Street datasets; the IWAE has only one house. Results are shown for different window sizes.

TABLE I
COMPARATIVE RESULTS ON REDD

| Window | MAE (kWh) | | | | | RMSE (kWh) | | | | | MAPE (%) | | | | |
|---|---|---|---|---|---|---|---|---|---|---|---|---|---|---|---|
| | ESN | LSTM | SC | RTL | $R^2TL$ | ESN | LSTM | SC | RTL | $R^2TL$ | ESN | LSTM | SC | RTL | $R^2TL$ |
| 2 days | 9.6 | 7.6 | 8.5 | 7.0 | 6.5 | 11.0 | 10.5 | 10.1 | 8.4 | 8.0 | 34.6 | 32.2 | 31.1 | 29.9 | 28.9 |
| 3 days | 9.0 | 6.6 | 8.0 | 6.2 | 5.4 | 9.4 | 8.7 | 9.1 | 8.0 | 7.6 | 31.3 | 30.4 | 30.2 | 28.7 | 27.5 |
| 4 days | 9.3 | 7.2 | 8.4 | 6.3 | 6.0 | 11.1 | 10.6 | 9.9 | 8.4 | 8.0 | 33.9 | 32.1 | 31.0 | 30.0 | 29.1 |
| 5 days | 10.7 | 8.1 | 9.9 | 7.4 | 6.7 | 12.6 | 11.8 | 11.2 | 9.7 | 8.8 | 34.4 | 33.8 | 32.6 | 31.0 | 30.3 |
| 6 days | 11.9 | 9.4 | 11.1 | 8.3 | 7.8 | 13.3 | 13.0 | 12.9 | 11.6 | 10.2 | 36.2 | 35.0 | 33.9 | 31.9 | 31.1 |
| 7 days | 13.4 | 11.8 | 12.7 | 9.4 | 8.7 | 15.8 | 15.0 | 13.5 | 12.6 | 11.5 | 37.5 | 36.7 | 35.3 | 33.2 | 32.0 |

TABLE II
COMPARATIVE RESULTS ON PECAN STREET

| Window | MAE (kWh) | | | | | RMSE (kWh) | | | | | MAPE (kWh) | | | | |
|---|---|---|---|---|---|---|---|---|---|---|---|---|---|---|---|
| | ESN | LSTM | SC | RTL | $R^2TL$ | ESN | LSTM | SC | RTL | $R^2TL$ | ESN | LSTM | SC | RTL | $R^2TL$ |
| 2 days | 7.9 | 6.6 | 5.4 | 5.0 | 4.4 | 9.0 | 8.3 | 6.5 | 6.2 | 5.6 | 32.7 | 31.2 | 25.1 | 23.9 | 22.0 |
| 3 days | 7.5 | 6.4 | 5.2 | 5.0 | 4.4 | 8.8 | 8.0 | 6.3 | 6.1 | 5.5 | 31.9 | 30.4 | 23.2 | 23.2 | 22.0 |
| 4 days | 7.2 | 6.3 | 5.2 | 4.7 | 4.2 | 8.4 | 7.5 | 6.3 | 6.0 | 5.5 | 30.9 | 29.8 | 23.1 | 23.0 | 21.5 |
| 5 days | 7.1 | 6.2 | 5.1 | 4.7 | 4.1 | 8.2 | 7.2 | 6.1 | 5.8 | 5.2 | 30.4 | 29.0 | 21.6 | 21.8 | 21.3 |
| 6 days | 7.1 | 6.2 | 5.1 | 4.6 | 4.0 | 8.2 | 7.0 | 6.0 | 5.8 | 5.2 | 30.2 | 28.4 | 21.5 | 21.7 | 21.3 |
| 7 days | 7.1 | 6.1 | 5.1 | 4.6 | 4.0 | 8.1 | 7.0 | 6.1 | 5.7 | 5.2 | 30.2 | 28.4 | 21.4 | 21.7 | 21.1 |

TABLE III
COMPARATIVE RESULTS ON IWAE

| Window | MAE (kWh) | | | | | RMSE (kWh) | | | | | MAPE (kWh) | | | | |
|---|---|---|---|---|---|---|---|---|---|---|---|---|---|---|---|
| | ESN | LSTM | SC | RTL | $R^2TL$ | ESN | LSTM | SC | RTL | $R^2TL$ | ESN | LSTM | SC | RTL | $R^2TL$ |

| | | | | | | | | | | | | | | |
|---|---|---|---|---|---|---|---|---|---|---|---|---|---|---|
| 2 days | 9.0 | 8.6 | 8.4 | 6.8 | 6.2 | 10.6 | 10.2 | 10.1 | 8.4 | 8.0 | 34.9 | 33.1 | 30.6 | 28.7 | 28.0 |
| 3 days | 8.5 | 7.9 | 7.9 | 6.1 | 5.6 | 9.8 | 9.3 | 9.1 | 8.0 | 7.6 | 32.0 | 30.8 | 30.0 | 28.0 | 27.0 |
| 4 days | 8.7 | 8.2 | 8.2 | 6.5 | 6.1 | 10.2 | 9.8 | 9.9 | 8.4 | 8.0 | 34.4 | 32.1 | 30.8 | 28.8 | 28.1 |
| 5 days | 9.0 | 8.7 | 9.5 | 7.1 | 6.6 | 11.5 | 10.4 | 11.2 | 9.7 | 8.8 | 35.8 | 33.6 | 32.0 | 30.1 | 28.9 |
| 6 days | 10.6 | 9.0 | 10.8 | 8.1 | 7.5 | 12.6 | 11.8 | 12.9 | 11.6 | 10.2 | 37.1 | 34.9 | 33.1 | 30.9 | 30.1 |
| 7 days | 11.9 | 10.5 | 12.0 | 9.2 | 8.3 | 14.3 | 13.7 | 13.5 | 12.6 | 11.5 | 38.6 | 36.6 | 34.2 | 32.4 | 31.5 |

From the tables I, II and III, we can infer that for smaller datasets such as REDD and IWAE, increasing size of the window does not always help improve the results. This is because, with the increase in the size of the window, the number of training samples decrease which in turn leads to over-fitting. For a large dataset like Pecan Street, increasing the window size improves the results as over-fitting is not pronounced. But the results tend to saturate after 5 days; indicating that the information about the remote past does not help improve the forecasting.

The second conclusion that we can draw from Tables I to III is that when the dataset is small, SC does not yield very good results. It is the worst, the results are almost the same as the baseline ESN. But for larger datasets, SC yields even better results than the state-of-the-art LSTM. This phenomenon can be attributed to over-fitting as well.

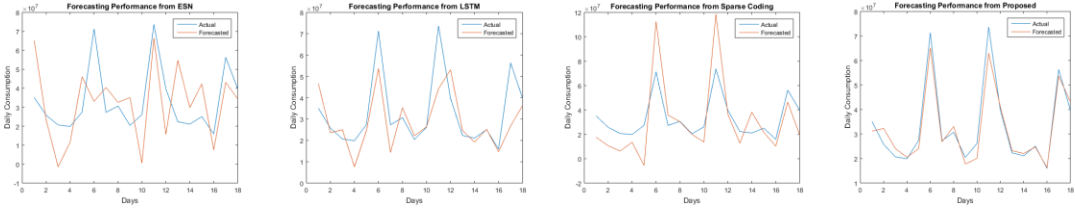

Fig. 4. Visualization of Forecasting Performance for REDD with 3 days window

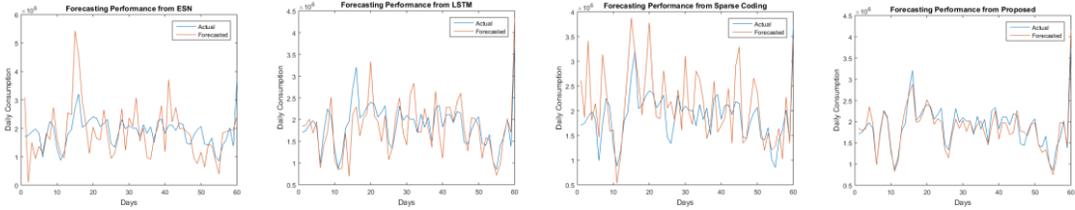

Fig. 5. Visualization of Forecasting Performance for Pecan Street with 5 days window

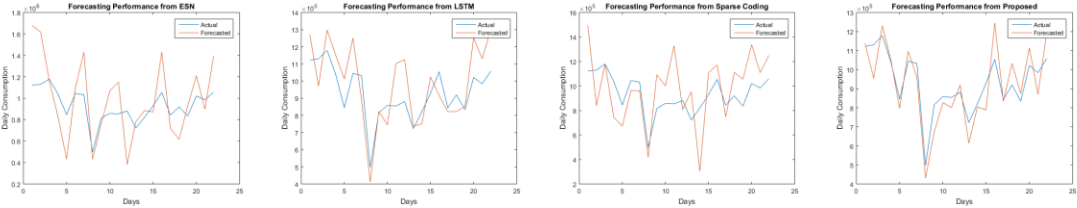

Fig. 6. Visualization of Forecasting Performance for IWAE with 3 days window

Whatever be the settings, our method yields the best results in terms of every possible metric. However, the $R^2TL$ formulation with in-built regression learning excels over the piecemeal RTL technique.

For visual comparison we show one randomly chosen house from the REDD and the Pecan Street dataset and the only house from the IWAE dataset. These are shown in Fig.s 4-6. We can easily see that our proposed method ($R^2TL$) is the only one that consistently follows the actual power consumption. The visual analysis corroborates the numerical results.

## IV. CONCLUSION

This work makes a fundamental contribution to dynamic modeling. So far dynamic modeling problems were restricted to primarily statistical models like auto-regressive moving average (ARMA) models and its variants, or linear / non-linear state space models like Kalman Filter and its non-linear variants. The other, relatively new, approach to dynamic modeling was based on reservoir computing in neural networks – recurrent neural network, echo state network and long short-term memory network. This work proposes a new approach to model dynamical systems based on the transform learning paradigm.

In this work, the technique has been used for the challenging problem of building level load forecasting. We show that our proposed technique outperforms all other techniques based on the models mentioned before.

The proposed technique is used here for load forecasting, but it is generic enough to be used to a variety of other problems in dynamical systems. Here we predicted a single value – load for the next day. In future, we would like to use it to predict load profile for an entire day or an entire week.

The technique developed here is generic enough to replace LSTM and other recurrent neural network models in any field of application; not necessarily restricted to dynamical systems. For example, these days LSTMs are used in vision problems. In principle our RTL technique can be replace LSTMs in such problems.

ACKNOWLEDGMENT

The first author is partially supported by the TCS PhD Research Fellowship. The second author is partially supported by the Infosys Center for Artificial Intelligence @ IIITD and by the DST IC-IMPACTS Indo-Canadian Grant.

APPENDIX

We have compared with the state-of-the-art techniques in load forecasting in the main manuscript. As mentioned before, we show the results for conventional approaches in this section. We have compared with the ARIMA, basic transform learning (TL) and the benchmark proposed by Hong et al in [45]. As can be seen, these methods are considerably worse than the state-of-the-art; hence we do not show them in the main manuscript.

TABLE IV
COMPARATIVE RESULTS ON REDD

| Window | MAE (kWh) | | | | | RMSE (kWh) | | | | | MAPE (%) | | | | |
|---|---|---|---|---|---|---|---|---|---|---|---|---|---|---|---|
| | ARIMA | TL | Hong | RTL | $R^2$TL | ARIMA | TL | Hong | RTL | $R^2$TL | ARIMA | TL | Hong | RTL | $R^2$TL |
| 2 days | 12.6 | 14.2 | 10.5 | 7.0 | 6.5 | 13.1 | 13.7 | 11.0 | 8.4 | 8.0 | 40.2 | 40.9 | 36.2 | 29.9 | 28.9 |
| 3 days | 12.1 | 13.6 | 10.2 | 6.2 | 5.4 | 12.4 | 12.7 | 10.1 | 8.0 | 7.6 | 35.2 | 35.4 | 34.5 | 28.7 | 27.5 |
| 4 days | 14.3 | 15.5 | 11.9 | 6.3 | 6.0 | 14.0 | 13.9 | 10.9 | 8.4 | 8.0 | 37.7 | 37.3 | 35.0 | 30.0 | 29.1 |
| 5 days | 16.2 | 16.3 | 12.8 | 7.4 | 6.7 | 15.2 | 14.9 | 11.4 | 9.7 | 8.8 | 39.8 | 38.8 | 36.8 | 31.0 | 30.3 |
| 6 days | 16.9 | 16.6 | 13.6 | 8.3 | 7.8 | 15.9 | 15.4 | 12.2 | 11.6 | 10.2 | 40.9 | 40.0 | 37.9 | 31.9 | 31.1 |
| 7 days | 17.5 | 17.0 | 13.7 | 9.4 | 8.7 | 16.3 | 16.0 | 12.8 | 12.6 | 11.5 | 41.5 | 40.7 | 38.6 | 33.2 | 32.0 |

TABLE V
COMPARATIVE RESULTS ON PECAN STREET

| Window | MAE (kWh) | | | | | RMSE (kWh) | | | | | MAPE (kWh) | | | | |
|---|---|---|---|---|---|---|---|---|---|---|---|---|---|---|---|
| | ARIMA | TL | Hong | RTL | $R^2$TL | ARIMA | TL | Hong | RTL | $R^2$TL | ARIMA | TL | Hong | RTL | $R^2$TL |
| 2 days | 9.8 | 9.9 | 8.7 | 5.0 | 4.4 | 12.1 | 12.5 | 9.6 | 6.2 | 5.6 | 36.7 | 36.7 | 35.1 | 23.9 | 22.0 |
| 3 days | 9.4 | 9.8 | 8.5 | 5.0 | 4.4 | 11.6 | 12.2 | 9.4 | 6.1 | 5.5 | 35.9 | 35.8 | 33.2 | 23.2 | 22.0 |
| 4 days | 9.3 | 9.5 | 8.3 | 4.7 | 4.2 | 11.0 | 11.5 | 9.3 | 6.0 | 5.5 | 34.9 | 34.0 | 32.1 | 23.0 | 21.5 |
| 5 days | 9.2 | 9.3 | 8.2 | 4.7 | 4.1 | 11.0 | 11.0 | 9.1 | 5.8 | 5.2 | 34.0 | 33.5 | 31.3 | 21.8 | 21.3 |
| 6 days | 9.2 | 9.2 | 8.2 | 4.6 | 4.0 | 10.8 | 10.7 | 9.0 | 5.8 | 5.2 | 33.2 | 33.2 | 30.5 | 21.7 | 21.3 |
| 7 days | 9.1 | 9.1 | 8.2 | 4.6 | 4.0 | 10.5 | 10.7 | 9.0 | 5.7 | 5.2 | 33.2 | 33.0 | 30.4 | 21.7 | 21.1 |

TABLE VI
COMPARATIVE RESULTS ON IWAE

| Window | MAE (kWh) | | | | | RMSE (kWh) | | | | | MAPE (kWh) | | | | |
|---|---|---|---|---|---|---|---|---|---|---|---|---|---|---|---|
| | ARIMA | TL | Hong | RTL | $R^2$TL | ARIMA | TL | Hong | RTL | $R^2$TL | ARIMA | TL | Hong | RTL | $R^2$TL |
| 2 days | 13.1 | 12.9 | 10.4 | 6.8 | 6.2 | 13.5 | 13.2 | 12.1 | 8.4 | 8.0 | 38.9 | 38.0 | 34.6 | 28.7 | 28.0 |
| 3 days | 12.6 | 12.0 | 9.8 | 6.1 | 5.6 | 12.8 | 13.0 | 11.1 | 8.0 | 7.6 | 37.6 | 37.8 | 33.0 | 28.0 | 27.0 |
| 4 days | 12.8 | 12.7 | 9.2 | 6.5 | 6.1 | 13.2 | 13.8 | 10.9 | 8.4 | 8.0 | 38.3 | 37.0 | 34.7 | 28.8 | 28.1 |
| 5 days | 13.0 | 12.9 | 10.1 | 7.1 | 6.6 | 13.5 | 14.4 | 11.8 | 9.7 | 8.8 | 39.1 | 38.7 | 36.1 | 30.1 | 28.9 |
| 6 days | 13.6 | 13.0 | 11.2 | 8.1 | 7.5 | 14.0 | 14.8 | 12.8 | 11.6 | 10.2 | 40.0 | 39.8 | 37.1 | 30.9 | 30.1 |
| 7 days | 13.9 | 13.4 | 12.0 | 9.2 | 8.3 | 14.5 | 14.9 | 13.8 | 12.6 | 11.5 | 41.1 | 40.5 | 38.3 | 32.4 | 31.5 |